\documentclass[journal]{IEEEtran}
\usepackage{cite}
\usepackage{amsmath,amssymb,amsfonts}
\usepackage{algorithmic}
\usepackage{graphicx}
\usepackage{textcomp}
\usepackage{xcolor}
\usepackage{fvextra}
\usepackage{listings}
\usepackage[T1]{fontenc}
\usepackage{lmodern} 
\usepackage{booktabs}
\usepackage[hidelinks]{hyperref}
\usepackage{subcaption}
\usepackage{amsmath}
\usepackage{amssymb}

\ifCLASSINFOpdf
\else
\fi
\begin{document}
%
\title{CyboRacket: A Perception-to-Action Framework for Humanoid Racket Sports}
%
%
%

\author{
\IEEEauthorblockN{
Peng Ren\textsuperscript{1,2,*}\thanks{Equal contribution.},
Chuan Qi\textsuperscript{3,2,*},
Haoyang Ge\textsuperscript{4,2,*},
Qiyuan Su\textsuperscript{5},
Xuguo He\textsuperscript{5},
Cong Huang\textsuperscript{2,6},
Pei Chi\textsuperscript{1,$\dagger$},
Jiang Zhao\textsuperscript{1},
Kai Chen\textsuperscript{2,5,6,$\dagger$},
}

\IEEEauthorblockA{
\textsuperscript{1}BUAA~~~
\textsuperscript{2}BZA~~~
\textsuperscript{3}USTC~~~
\textsuperscript{4}TJU~~~
\textsuperscript{5}DeepCybo~~~
\textsuperscript{6}ZGCI\\
\textsuperscript{*}Equal contribution.\\
\textsuperscript{$\dagger$}Corresponding authors: Pei Chi (peichi@buaa.edu.cn), Kai Chen (kaichen@zgci.ac.cn).
}
}

%
%

\markboth{Journal of \LaTeX\ Class Files,~Vol.~14, No.~8, August~2015}%
{Shell \MakeLowercase{\textit{et al.}}: Bare Demo of IEEEtran.cls for IEEE Journals}
%



\maketitle

\begin{abstract}
Dynamic ball-interaction tasks remain challenging for robots because they require tight perception-action coupling under limited reaction time. This challenge is especially pronounced in humanoid racket sports, where successful interception depends on accurate visual tracking, trajectory prediction, coordinated stepping, and stable whole-body striking. Existing robotic racket-sport systems often rely on external motion capture for state estimation or on task-specific low-level controllers that must be retrained across tasks and platforms. We present \emph{CyboRacket}, a hierarchical perception-to-action framework for humanoid racket sports that integrates onboard visual perception, physics-based trajectory prediction, and large-scale pre-trained whole-body control. The framework uses onboard cameras to track the incoming object, predicts its future trajectory, and converts the estimated interception state into target end-effector and base-motion commands for whole-body execution by SONIC on the Unitree G1 humanoid robot. We evaluate the proposed framework in a vision-based humanoid tennis-hitting task. Experimental results demonstrate real-time visual tracking, trajectory prediction, and successful striking using purely onboard sensing.
\end{abstract}


%
\IEEEpeerreviewmaketitle

\section{Introduction}
%
%
%
%

\IEEEPARstart{R}{ecent} years have seen substantial progress in robotic control across diverse platforms, from quadrupeds traversing challenging terrain to industrial manipulators performing precise assembly and humanoids achieving stable locomotion and whole-body coordination~\cite{long2025learning,zhang2025hub,cheng2024expressive}. However, tasks involving fast-moving objects remain significantly more challenging, as they require tight perception-action coupling under limited reaction time. Robotic racket sports, including table tennis, badminton, and tennis, exemplify such dynamic visuomotor problems: the robot must track a high-speed ball, predict its future trajectory, and generate coordinated striking motions with accurate spatial and temporal alignment.

Existing approaches to robotic racket-sport interaction can be broadly divided into two categories. End-to-end learning methods map sensory observations directly to low-level actions without explicit intermediate representations. For example, Tebbe et al. showed that reinforcement learning can acquire rallying behavior from scratch in fewer than 200 trials~\cite{tebbe2021sample}, while the Gemini Robotics team demonstrated amateur human-level robotic table tennis through large-scale hierarchical reinforcement learning and sim-to-real transfer~\cite{dambrosio2023robotic,dambrosio2025achieving}. In contrast, hierarchical methods separate projectile-state prediction and interception planning from low-level motion execution, typically combining model-based trajectory prediction and strike-point generation with specialized controllers such as MPC~\cite{nguyen2025high} or task-specific learned policies~\cite{su2025hitter,ma2025learning}. While more modular and interpretable, existing hierarchical systems still face two key limitations: many depend on external motion capture or other instrumented sensing for accurate state estimation~\cite{su2025hitter}, and their low-level controllers are often task-specific, requiring retraining when the task or platform changes~\cite{ma2025learning}. This motivates a hierarchical framework that integrates onboard closed-loop perception with a reusable pre-trained whole-body execution backbone.

Among these dynamic racket-sport settings, humanoid platforms are particularly challenging. Unlike fixed-base manipulators, humanoid robots must coordinate perception, locomotion, balance maintenance, and arm swinging within a single whole-body control loop. The reachable striking region is not determined by the arm alone, but also depends on timely base repositioning and stable lower-body support. As a result, errors in state estimation or interception timing can propagate not only to racket contact, but also to stepping behavior and whole-body balance. These characteristics make humanoid racket sports a demanding testbed for dynamic perception-action integration.

In this work, we present \emph{CyboRacket}, a hierarchical perception-to-action framework for humanoid racket sports that integrates onboard visual perception, physics-based trajectory prediction, and large-scale pre-trained whole-body control. The system uses a YOLO-based detector running entirely on onboard cameras to track fast incoming projectiles in real time, without relying on external motion capture. A physics-based predictor estimates the future trajectory, including the landing region and arrival time, and converts these predictions into target end-effector states and base motion commands for interception. To realize these commands on a humanoid platform, we build the execution layer on top of SONIC~\cite{luo2025sonic}, a recently open-sourced large-scale motion tracking foundation model for humanoid control. We instantiate the proposed framework on the Unitree G1 humanoid robot in a vision-based tennis-hitting task. By coupling closed-loop perception and interception planning with pre-trained whole-body motion execution, the framework enables time-critical projectile interaction through coordinated stepping and swinging. Experimental results show that the proposed system supports real-time projectile tracking, trajectory prediction, and successful striking using purely onboard visual sensing, demonstrating the feasibility of perception-driven dynamic racket-sport interaction on a humanoid platform.

The main contributions of this work are as follows:
\begin{itemize}
    \item We propose a hierarchical perception-to-action framework for humanoid racket sports that combines onboard visual trajectory prediction with a large-scale pre-trained humanoid motion executor, avoiding task-specific retraining of the full low-level controller.
    \item We develop a purely onboard visual closed-loop pipeline for fast-projectile perception and trajectory prediction, removing the dependence on external motion capture systems.
    \item We demonstrate vision-based tennis hitting on the Unitree G1 humanoid robot, showing the effectiveness of integrating pre-trained whole-body motion control with closed-loop perception and planning for dynamic humanoid interaction.
\end{itemize}

\section{Related Work}

\subsection{Robotic Racket Sports and Dynamic Object Interaction}

Robotic racket sports provide a representative benchmark for dynamic perception, prediction, and time-critical control. Early table-tennis systems established the foundations for high-speed striking through biologically inspired trajectory generation, kinesthetic teach-in, and anticipatory action selection~\cite{mulling2010biomimetic,mulling2013learning,wang2011learning,wang2017anticipatory}. More recent work has increasingly adopted learning-based methods. The Gemini Robotics team demonstrated amateur human-level robotic table tennis using hierarchical learning with sim-to-real transfer~\cite{dambrosio2023robotic,dambrosio2025achieving}, while Tebbe et al. showed that reinforcement learning can acquire rallying behavior from scratch in fewer than 200 trials~\cite{tebbe2021sample}.

Related developments in other racket-sport and dynamic-manipulation settings have further extended these ideas beyond fixed-base platforms. Zaidi et al. developed an athletic mobile manipulator for wheelchair tennis~\cite{zaidi2023athletic}, and Ma et al. demonstrated coordinated badminton skills on legged manipulators with onboard perception and whole-body visuomotor coordination~\cite{ma2025learning}. On humanoid platforms, Xiong et al. demonstrated table tennis with impedance control but were limited to static standing~\cite{xiong2012impedance}, whereas Su et al. combined model-based planning with RL-based whole-body control to enable dynamic humanoid table tennis~\cite{su2025hitter}. Despite this progress, current systems still exhibit two recurring limitations: many depend on external motion capture or other privileged sensing setups for accurate projectile-state estimation~\cite{su2025hitter}, and their control policies are often specialized for a specific task or platform~\cite{ma2025learning}. These limitations make it difficult to achieve deployable onboard perception together with reusable whole-body execution for humanoid racket-sport tasks.

\subsection{Humanoid Whole-Body Control and Scalable Motion Tracking}

In parallel, humanoid control has evolved from task-specific imitation toward scalable whole-body motion tracking. Early work focused on learning controllers for individual motions or motion families through large-scale reinforcement learning and manually designed rewards~\cite{peng2018deepmimic,peng2021amp}. Subsequent approaches improved coordination by decoupling upper- and lower-body control~\cite{he2024omnih2o,zhang2025falcon}, while more recent methods developed general whole-body motion trackers capable of reproducing a broad range of human movements~\cite{liao2025beyondmimic,chen2025gmt}.

A growing line of work has further explored large-scale pre-trained models as reusable control backbones for humanoid robots. Luo et al. showed that scaling motion tracking to 100 million frames can yield universal tracking capabilities across diverse behaviors, suggesting motion tracking as a practical foundation for humanoid control~\cite{luo2025sonic}. SONIC~\cite{luo2025sonic} and contemporaneous systems~\cite{ze2025twist,zhang2025track} further demonstrated that such pre-trained controllers can support multimodal command interfaces, including teleoperation, video-driven control, and integration with higher-level vision-language-action models~\cite{bjorck2025groot}. However, these methods mainly focus on tracking externally provided motion references and remain weakly connected to task-level closed-loop perception. As a result, although they provide strong reusable whole-body motion priors, they do not by themselves solve dynamic interactive tasks such as racket sports, where perception, prediction, and action must be tightly coupled in real time.

\section{System Overview}

\begin{figure*}[t]
  \centering
  \includegraphics[width=0.9\textwidth]{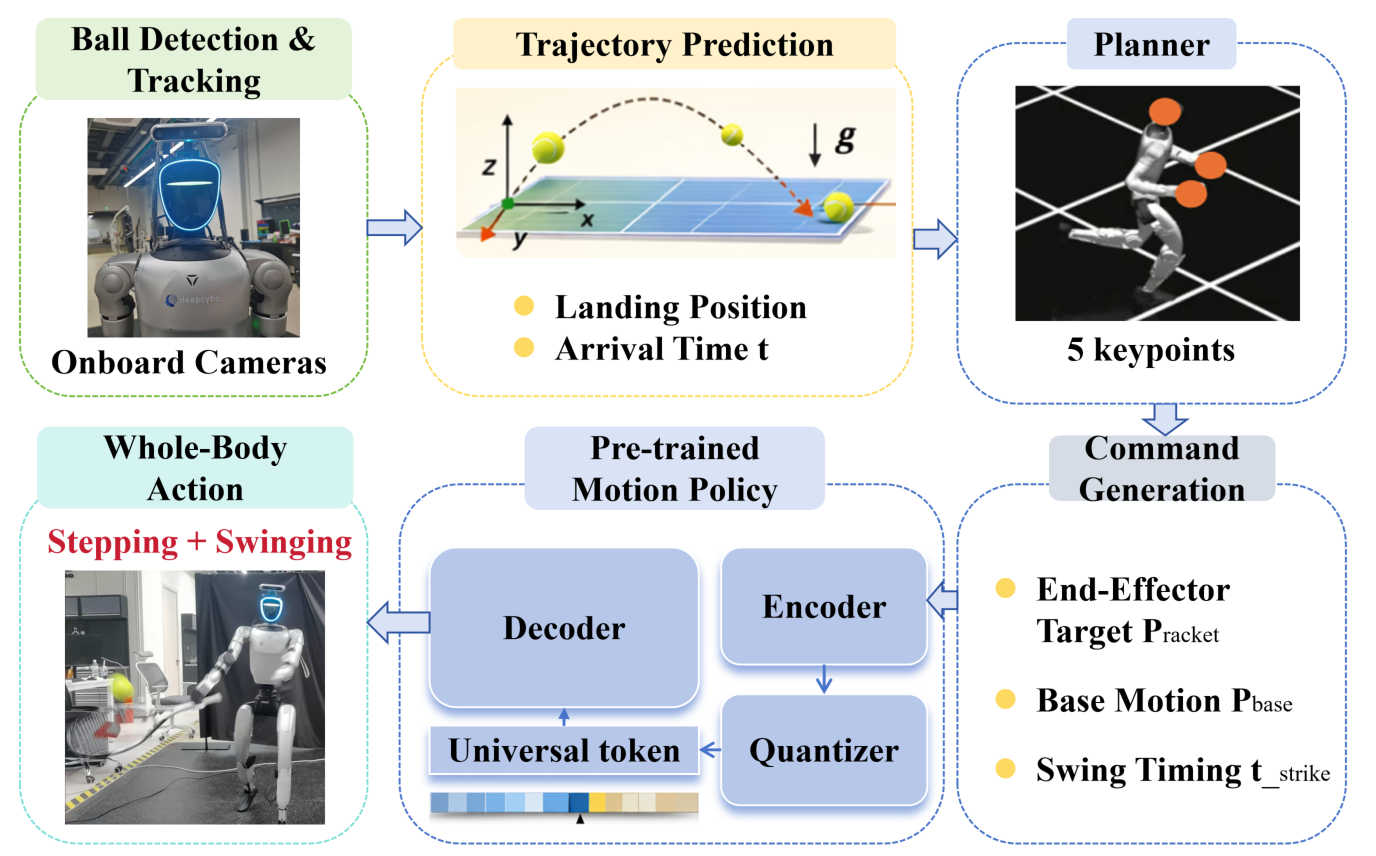}
  \caption{System overview of the humanoid  humanoid racket sports framework.}
  \label{fig:system}
\end{figure*}

We address real-time tennis interception on the Unitree G1 humanoid as an onboard perception-to-action problem, where noisy RGB-D observations of a fast incoming projectile must be converted into dynamically feasible whole-body striking behaviors under tight latency constraints. As shown in Fig.~\ref{fig:system}, our framework closes this loop through three tightly coupled stages: projectile state estimation from a head-mounted RGB-D camera, interception target generation from short-horizon trajectory prediction, and whole-body execution through a SONIC-based control interface. The robot is equipped with a 100~Hz RGB-D camera mounted on the head, whose synchronized color and depth streams are processed together with proprioceptive robot states for visual tracking, geometric reconstruction, and frame transformation. From image-space projectile observations, the system maintains temporally consistent 2D measurements, reconstructs 3D projectile positions in the robot frame, and estimates a filtered projectile state using an EKF. This state estimator further rolls out the short-horizon trajectory of the projectile and continuously provides the downstream planner with updated interception candidates.

Given the estimated projectile trajectory, the planner selects a dynamically feasible interception target by jointly considering timing, reachability, body stability, and swing effectiveness. The planning process runs online in a receding-horizon manner and is coupled with a swing state machine, allowing the robot to revise its interception decision as new visual evidence arrives. Instead of directly producing joint-level commands, the planner outputs structured motion targets, including the desired head orientation for maintaining visual continuity, wrist targets for racket interception, body height adjustment, and navigation velocity. These targets are mapped to the low-dimensional SONIC command interface, which drives coordinated locomotion and striking behaviors at the whole-body level. This design separates fast visual updates from motion generation while preserving a tight closed loop between onboard perception, interception planning, and physically executable humanoid control.

\section{Onboard RGB-D projectile Perception and State Estimation}

This section presents an onboard RGB-D perception pipeline that estimates the state of the incoming racket-sport projectile and provides the planner with a temporally consistent 3D trajectory prediction. In the current implementation, we instantiate this module for tennis-ball perception. The main challenge is that the projectile appears as a small, fast-moving target under limited sensing resolution and short reaction horizons, while aggressive head and body motions can further amplify localization noise and timing error. Motivated by the closed-loop requirements of dynamic interception, we formulate this perception module not as isolated detection and tracking components, but as a unified estimation process that converts synchronized RGB-D measurements into uncertainty-aware 3D projectile states and short-horizon trajectory predictions.

\subsection{Projectile Observation and 2D Tracking}

We first extract image-space projectile observations from synchronized RGB and depth frames captured by the head-mounted RGB-D camera. A YOLO-based detector is applied to the RGB image to produce candidate tennis-ball bounding boxes. To preserve geometric consistency between the network input and the original image coordinates, the RGB image is resized with aspect-ratio-preserving padding before inference, and the predicted boxes are mapped back to the original image plane afterward. Since generic \texttt{sports ball} detections are prone to false positives in real scenes, we further apply task-specific filtering based on geometry and appearance. In particular, for the tennis instantiation used in this work, we reject candidates with implausible scale or aspect ratio and retain only regions whose color statistics are consistent with the typical yellow-green appearance of a tennis ball. The resulting output is a timestamped 2D observation stream containing the projectile bounding box, confidence, and image metadata.

To improve robustness against single-frame errors, missed detections, and motion blur, we maintain a lightweight single-object temporal tracker in image space. The tracker propagates the current observation with a constant-velocity model and associates new detections using center proximity, scale consistency, and motion-direction agreement. This produces a temporally smoothed 2D projectile observation and also serves as an observation manager: when detections are briefly unavailable, the tracker extrapolates the object center for a small number of frames; when the observation gap exceeds a threshold, the trajectory is reset. Compared with raw frame-wise detections, this representation provides a substantially more stable visual signal for downstream 3D reconstruction and state estimation.

\subsection{RGB-D 3D Localization and Measurement Uncertainty}

The tracked 2D observation is then lifted to 3D using the aligned depth measurement from the RGB-D camera. Rather than directly querying depth at a single pixel, we estimate the projectile depth from a local region centered at the tracked image location, which is more robust for small distant objects whose depth measurements are easily corrupted by background leakage, invalid samples, and boundary noise. Invalid or out-of-range values are removed, and a foreground-biased robust statistic is used to recover the depth value.

Let $(u_t,v_t)$ denote the tracked  projectile  center in the RGB image at time $t$, and let $Z_t$ denote the aggregated depth value obtained from the aligned depth map within a local region around the  projectile  center. Given the RGB-D camera intrinsics
\[
\mathbf{K}=
\begin{bmatrix}
f_x&0&c_x\\
0&f_y&c_y\\
0&0&1
\end{bmatrix},
\]
where $(f_x,f_y)$ are the focal lengths and $(c_x,c_y)$ is the principal point, the 3D projectile position in the camera frame is recovered by pinhole back-projection as
\begin{equation}
X_t = \frac{(u_t-c_x)Z_t}{f_x},\qquad
Y_t = \frac{(v_t-c_y)Z_t}{f_y},\qquad
\mathbf{p}_{c,t}=
\begin{bmatrix}
X_t\\Y_t\\Z_t
\end{bmatrix}.
\label{eq:backprojection}
\end{equation}
The point is then transformed into the robot base frame as
\begin{equation}
\mathbf{p}_{b,t} = \mathbf{R}_{b\leftarrow c}\mathbf{p}_{c,t} + \mathbf{t}_{b\leftarrow c},
\label{eq:camera_to_base}
\end{equation}
where $\mathbf{R}_{b\leftarrow c}\in SO(3)$ and $\mathbf{t}_{b\leftarrow c}\in\mathbb{R}^3$ are the calibrated extrinsic rotation and translation from the camera frame to the robot base frame.

To quantify the reliability of each 3D observation, we further associate the reconstructed point with a measurement covariance matrix $\mathbf{R}^{\mathrm{meas}}_t$. We model uncertainty in the image-plane localization and depth estimate by
\begin{equation}
\Sigma_{uvZ,t}=
\mathrm{diag}\!\left(\sigma_{u,t}^2,\sigma_{v,t}^2,\sigma_{Z,t}^2\right),
\label{eq:uvz_cov}
\end{equation}
where $\sigma_{u,t}$ and $\sigma_{v,t}$ denote the standard deviations of the projectile center in pixel coordinates, and $\sigma_{Z,t}$ denotes the standard deviation of the aggregated depth estimate. We then propagate this uncertainty through the nonlinear mapping from $(u_t,v_t,Z_t)$ to the base-frame position $\mathbf{p}_{b,t}$ using first-order error propagation:
\begin{equation}
\mathbf{R}^{\mathrm{meas}}_t \approx \mathbf{J}_t\,\Sigma_{uvZ,t}\,\mathbf{J}_t^\top,
\qquad
\mathbf{J}_t=\frac{\partial \mathbf{p}_{b,t}}{\partial (u_t,v_t,Z_t)} .
\label{eq:meas_cov}
\end{equation}
Here, $\mathbf{J}_t\in\mathbb{R}^{3\times 3}$ denotes the Jacobian of the mapping from $(u_t,v_t,Z_t)$ to the reconstructed base-frame position $\mathbf{p}_{b,t}$, and thus characterizes how image-plane and depth uncertainties are amplified in 3D space. This construction yields a range-dependent measurement covariance: as the projectile moves farther away, a given pixel or depth error induces a larger 3D localization error. The resulting covariance is used in the EKF update step to weight each observation according to its geometric reliability.

\subsection{EKF-Based projectile State Estimation}

On top of these reconstructed 3D measurements, we maintain a continuous estimate of the projectile state using an extended Kalman filter (EKF). The EKF state consists of the 3D projectile position and velocity,
\begin{equation}
\mathbf{x}_t = [\mathbf{p}_t^\top,\mathbf{v}_t^\top]^\top
= [p_{x,t},\, p_{y,t},\, p_{z,t},\, v_{x,t},\, v_{y,t},\, v_{z,t}]^\top,
\label{eq:ekf_state}
\end{equation}
where $\mathbf{p}_t=[p_{x,t},p_{y,t},p_{z,t}]^\top$ denotes the projectile position and $\mathbf{v}_t=[v_{x,t},v_{y,t},v_{z,t}]^\top$ denotes the projectile velocity in the robot base frame. The measurement vector contains only the reconstructed 3D position,
\begin{equation}
\mathbf{z}_t = \mathbf{H}\mathbf{x}_t + \mathbf{n}_t,\qquad
\mathbf{H}=[\mathbf{I}_3\ \mathbf{0}_{3\times 3}],\qquad
\mathbf{n}_t\sim\mathcal{N}(\mathbf{0},\mathbf{R}^{\mathrm{meas}}_t),
\label{eq:ekf_measurement}
\end{equation}
where $\mathbf{H}$ extracts the position component from the full state and $\mathbf{R}^{\mathrm{meas}}_t$ is the measurement covariance defined in (\ref{eq:meas_cov}).

During the prediction step, the projectile state is propagated with a discrete flight model that accounts for gravity and aerodynamic drag. Starting from the current state estimate $(\mathbf{p}_t,\mathbf{v}_t)$ and time step $\Delta t$, we update the position and velocity as
\begin{equation}
\mathbf{p}_{t+1} = \mathbf{p}_t + \mathbf{v}_t \Delta t + \frac{1}{2}\mathbf{a}(\mathbf{v}_t)\Delta t^2,
\qquad
\mathbf{v}_{t+1} = \mathbf{v}_t + \mathbf{a}(\mathbf{v}_t)\Delta t,
\label{eq:ekf_dynamics}
\end{equation}
where the acceleration term is defined as
\begin{equation}
\mathbf{a}(\mathbf{v}) = \mathbf{g} - k_1 \mathbf{v} - k_2 \|\mathbf{v}\|\mathbf{v},
\qquad
\mathbf{g} = [0,0,-g]^\top,
\label{eq:ball_acc}
\end{equation}
with $\mathbf{g}$ the gravitational acceleration and $k_1,k_2$ the coefficients of linear and quadratic drag, respectively. In compact form, the EKF process model is written as
\begin{equation}
\mathbf{x}_{t+1} = f(\mathbf{x}_t,\Delta t) + \mathbf{w}_t,
\qquad
\mathbf{w}_t \sim \mathcal{N}(\mathbf{0},\mathbf{Q}_t),
\label{eq:ekf_process}
\end{equation}
where $\mathbf{w}_t$ models process uncertainty and $\mathbf{Q}_t$ is the corresponding process covariance.

During the measurement update, we first validate each incoming 3D observation using a Mahalanobis-distance gate on the innovation, in order to reject outliers caused by false detections, depth corruption, or abrupt mismatches. The innovation is defined as
\begin{equation}
\mathbf{r}_t = \mathbf{z}_t - \mathbf{H}\hat{\mathbf{x}}_{t|t-1},
\label{eq:innovation}
\end{equation}
where $\hat{\mathbf{x}}_{t|t-1}$ is the predicted EKF state and $\mathbf{H}$ is the measurement matrix defined in (\ref{eq:ekf_measurement}). The corresponding innovation covariance is
\begin{equation}
\mathbf{S}_t = \mathbf{H}\mathbf{P}_{t|t-1}\mathbf{H}^\top + \mathbf{R}^{\mathrm{meas}}_t,
\label{eq:innovation_cov}
\end{equation}
with $\mathbf{P}_{t|t-1}$ the predicted state covariance. We then compute the Mahalanobis distance
\begin{equation}
d_t^2 = \mathbf{r}_t^\top \mathbf{S}_t^{-1} \mathbf{r}_t,
\label{eq:mahalanobis}
\end{equation}
and accept the measurement only if $d_t^2 < \tau$, where $\tau$ is chosen according to a $\chi^2$ threshold. Otherwise, the update is skipped and the filter continues prediction-only propagation, which allows the estimator to bridge short observation gaps without being corrupted by unreliable measurements.

To better capture near-ground projectile motion, we additionally apply a simplified bounce update when the estimated projectile height in the robot base frame falls below a bounce threshold $z_b$ and the vertical velocity is downward, i.e., $\hat{p}_z<z_b$ and $\hat{v}_z<0$. In this case, the post-bounce vertical velocity is approximated by
\begin{equation}
v_z^{+} = -e\,v_z^{-},
\qquad e\in(0,1),
\label{eq:bounce}
\end{equation}
where $v_z^{-}$ and $v_z^{+}$ denote the pre- and post-bounce vertical velocities, respectively, and $e$ is an effective restitution coefficient. Optionally, the corresponding covariance can be inflated after the bounce event to reflect additional uncertainty introduced by ground contact. This approximation is sufficient for short-horizon interception prediction while avoiding the complexity of a more detailed contact model.

\subsection{Short-Horizon Trajectory Prediction}

Beyond recursive filtering, we further perform short-horizon trajectory prediction for interception planning. Starting from the current filtered state estimate, we roll out the same projectile dynamics model forward in time by discrete integration and obtain a sequence of future trajectory samples
\begin{equation}
\hat{\mathcal{T}} = \{\hat{\mathbf{p}}(\tau_i),\,\hat{\mathbf{v}}(\tau_i)\}_{i=1}^{N},
\label{eq:traj_rollout}
\end{equation}
where $\tau_i=i\Delta t_{\mathrm{pred}}$ denotes the prediction horizon. From this rollout, we compute interception-relevant quantities, including the predicted hitting point $\mathbf{p}_{\mathrm{hit}}$, the corresponding hitting time $t_{\mathrm{hit}}$, the projectile velocity at interception $\mathbf{v}_{\mathrm{hit}}$, and the predicted landing point.

In particular, the candidate hitting point is determined as the earliest feasible intersection between the predicted trajectory and a designated hitting-height plane within the prediction horizon,
\begin{equation}
\hat{p}_z(t_{\mathrm{hit}})=z_0,
\label{eq:hit_plane}
\end{equation}
where $z_0$ is the desired hitting height. A candidate interception state is exported to the planner only when the EKF state covariance remains below a threshold, the predicted hitting time lies within a feasible reaction window, and the target remains inside the robot's reachable workspace. In this way, the estimator provides the planner not only with a filtered instantaneous projectile state but also with a dynamically consistent short-horizon forecast for online interception.

\section{Interception Planning and Pre-trained Whole-Body Control}

This section converts the estimated projectile state into executable whole-body interception commands. Rather than synthesizing joint-space trajectories directly, the planner outputs a structured interception intent, including the desired contact time and location, target base placement, swing direction, and execution phase, while locomotion realization, balance maintenance, and coordinated whole-body motion are delegated to the pre-trained SONIC execution stack. This decomposition keeps the planning problem compact and interpretable while leveraging SONIC as a scalable whole-body control backbone.

\subsection{Interception Planning}

At each planning cycle, the planner receives the filtered projectile state, a short-horizon trajectory prediction, and the current robot base pose. Let
\[
\hat{\mathcal{T}}=\{\hat{\mathbf{p}}(\tau_i),\hat{\mathbf{v}}(\tau_i)\}_{i=1}^{N}
\]
denote the predicted projectile trajectory sampled at future time instants $\tau_i$, where $\hat{\mathbf{p}}(\tau_i)\in\mathbb{R}^3$ and $\hat{\mathbf{v}}(\tau_i)\in\mathbb{R}^3$ are the predicted position and velocity, and $N$ is the number of prediction samples. The planner first identifies a candidate interception state
$(\mathbf{p}_{\mathrm{hit}},\, t_{\mathrm{hit}})$
from the earliest feasible intersection between the predicted trajectory and the designated hitting-height plane, subject to predefined temporal and workspace constraints.

Given the candidate contact point, we compute a desired base placement from a nominal racket-contact template. Let $\mathbf{r}_{\mathrm{hit}}\in\mathbb{R}^3$ denote the preferred offset from the robot base to the racket contact point at impact, expressed in the base-heading frame. Given a desired base yaw $\psi_{\mathrm{des}}$, the corresponding target base position is
\begin{equation}
\mathbf{p}_{\mathrm{base}}^{\mathrm{des}}
=
\mathbf{p}_{\mathrm{hit}}
-
\mathbf{R}_z(\psi_{\mathrm{des}})\mathbf{r}_{\mathrm{hit}},
\label{eq:base_des}
\end{equation}
where $\mathbf{R}_z(\psi_{\mathrm{des}})\in SO(3)$ denotes the planar rotation matrix about the vertical axis. Intuitively, the planner selects a base pose such that the nominal racket contact point aligns with the predicted projectile position at the desired contact time.

The resulting interception plan at planning cycle $t$ is represented as
\begin{equation}
\Pi_t=
\bigl(
t_{\mathrm{hit}},
\mathbf{p}_{\mathrm{hit}},
\mathbf{p}_{\mathrm{base}}^{\mathrm{des}},
\psi_{\mathrm{des}},
\mathbf{d}_{\mathrm{swing}},
m_t,
\mathrm{valid}
\bigr),
\label{eq:intercept_plan}
\end{equation}
where $\mathbf{d}_{\mathrm{swing}}\in\mathbb{R}^3$ denotes the desired swing direction, $m_t \in \{\texttt{APPROACH},\texttt{SWING},\texttt{RECOVER}\}$ denotes the current swing mode, and $\mathrm{valid}\in\{0,1\}$ indicates whether the candidate is dynamically feasible. In practice, feasibility is checked through simple reachability and timing tests: the target is accepted only if the robot can translate and rotate its base into the desired pose within the available reaction time and if the predicted contact lies inside the arm-reachable hitting region. This formulation avoids full-body trajectory search and instead reduces interception planning to geometric reasoning over the predicted projectile trajectory and the reachable contact manifold.

\subsection{Online Replanning and Swing Coordination}

To maintain responsiveness under fast projectile motion, the overall system is organized as three asynchronous loops running at different rates. The \emph{perception loop} continuously updates the latest projectile-state estimate and handles short observation dropouts. The \emph{planning loop} periodically invokes the interception planner in~(\ref{eq:intercept_plan}) and maintains both the current valid plan and the most recent valid fallback plan. The \emph{command loop} runs at the highest frequency and converts the active plan into SONIC-compatible motion targets. This multi-rate design decouples fast state refresh from more expensive replanning while preserving smooth command output to the controller.

On top of this replanning mechanism, we maintain a swing state machine with three stages:
\[
\texttt{APPROACH}\rightarrow\texttt{SWING}\rightarrow\texttt{RECOVER}.
\]
During \texttt{APPROACH}, the robot moves toward the desired base pose while keeping the racket arm in a ready configuration and the head oriented toward the incoming projectile. Once the robot is sufficiently close to the target base pose and the predicted contact time enters a narrow execution window, the planner commits to \texttt{SWING}. In this stage, the racket arm follows a staged impact template consisting of a backswing (\emph{cock}), contact, and follow-through phase. After the predicted impact window has passed, the state machine transitions to \texttt{RECOVER}, where the upper body and head are smoothly returned toward a neutral ready posture for the next interception cycle.

This design separates \emph{when} to swing from \emph{how} the full body realizes the motion. The planner reasons over contact timing, base placement, and phase transitions, while the low-level realization of locomotion, balance, and coordinated whole-body motion is handled by SONIC.

\begin{figure*}[t]
  \centering
  \begin{subfigure}[t]{0.24\textwidth}
    \centering
    \includegraphics[width=\linewidth,height=3.5cm]{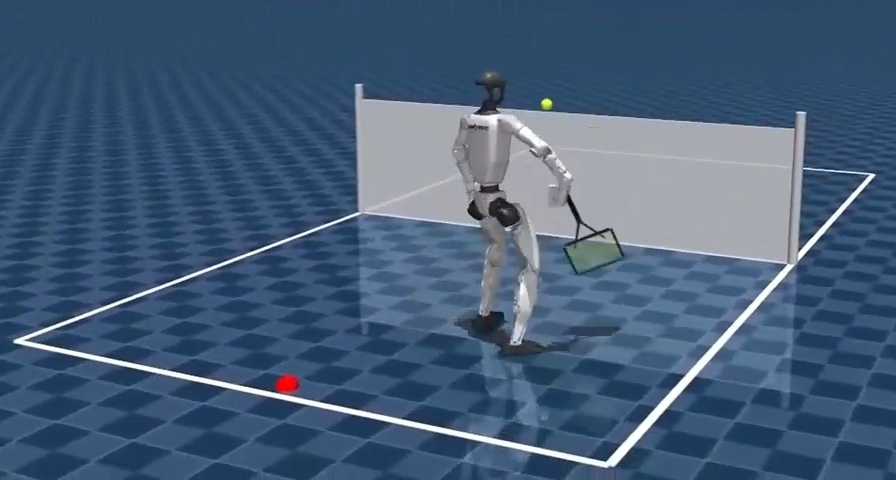}
  \end{subfigure}\hfill
  \begin{subfigure}[t]{0.24\textwidth}
    \centering
    \includegraphics[width=\linewidth,height=3.5cm]{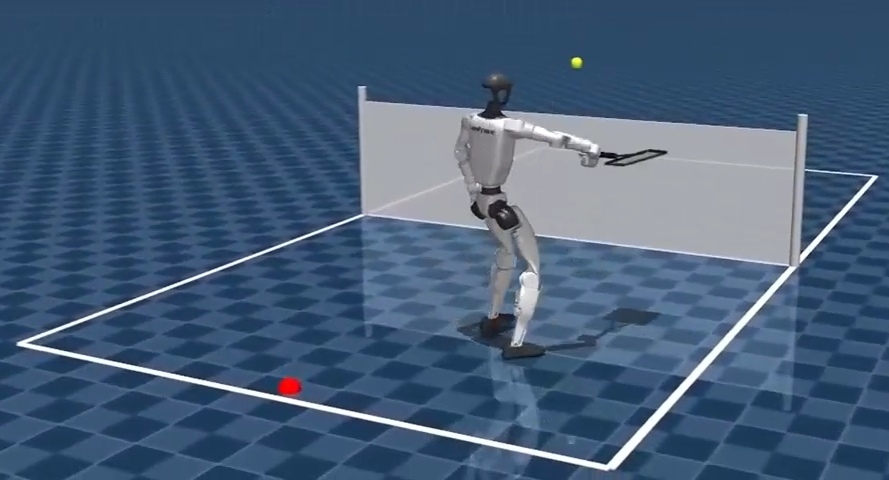}
  \end{subfigure}\hfill
  \begin{subfigure}[t]{0.24\textwidth}
    \centering
    \includegraphics[width=\linewidth,height=3.5cm]{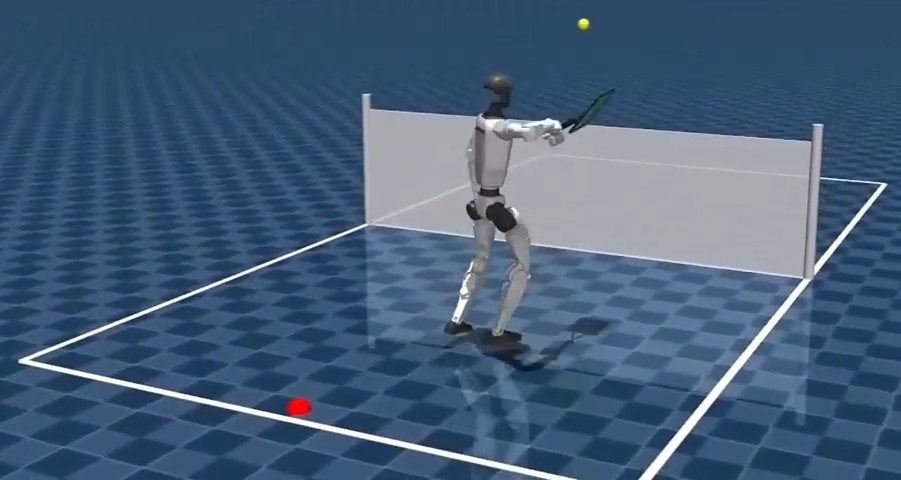}
  \end{subfigure}\hfill
  \begin{subfigure}[t]{0.24\textwidth}
    \centering
    \includegraphics[width=\linewidth,height=3.5cm]{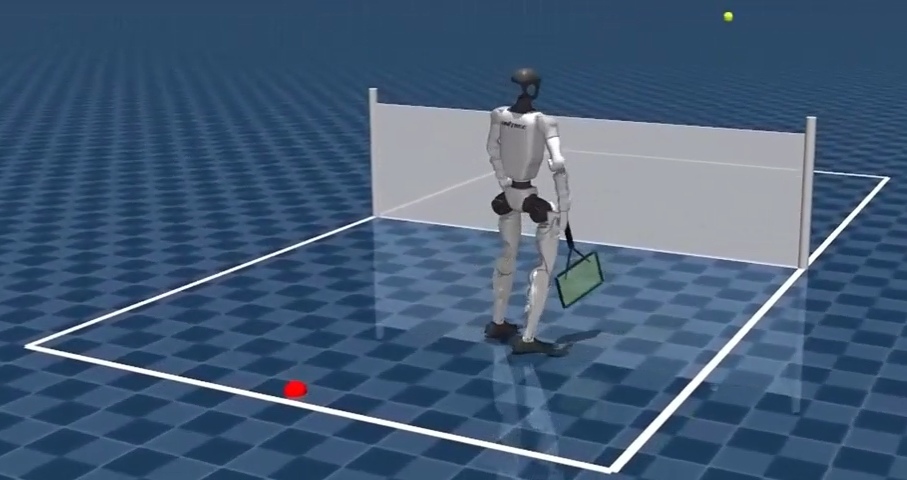}
  \end{subfigure}
  \caption{Representative motion sequence of humanoid badminton hitting in simulation.}
  \label{fig:s}
\end{figure*}

\begin{figure*}[t]
  \centering
  \begin{subfigure}[t]{0.24\textwidth}
    \centering
    \includegraphics[width=\linewidth,height=3.5cm]{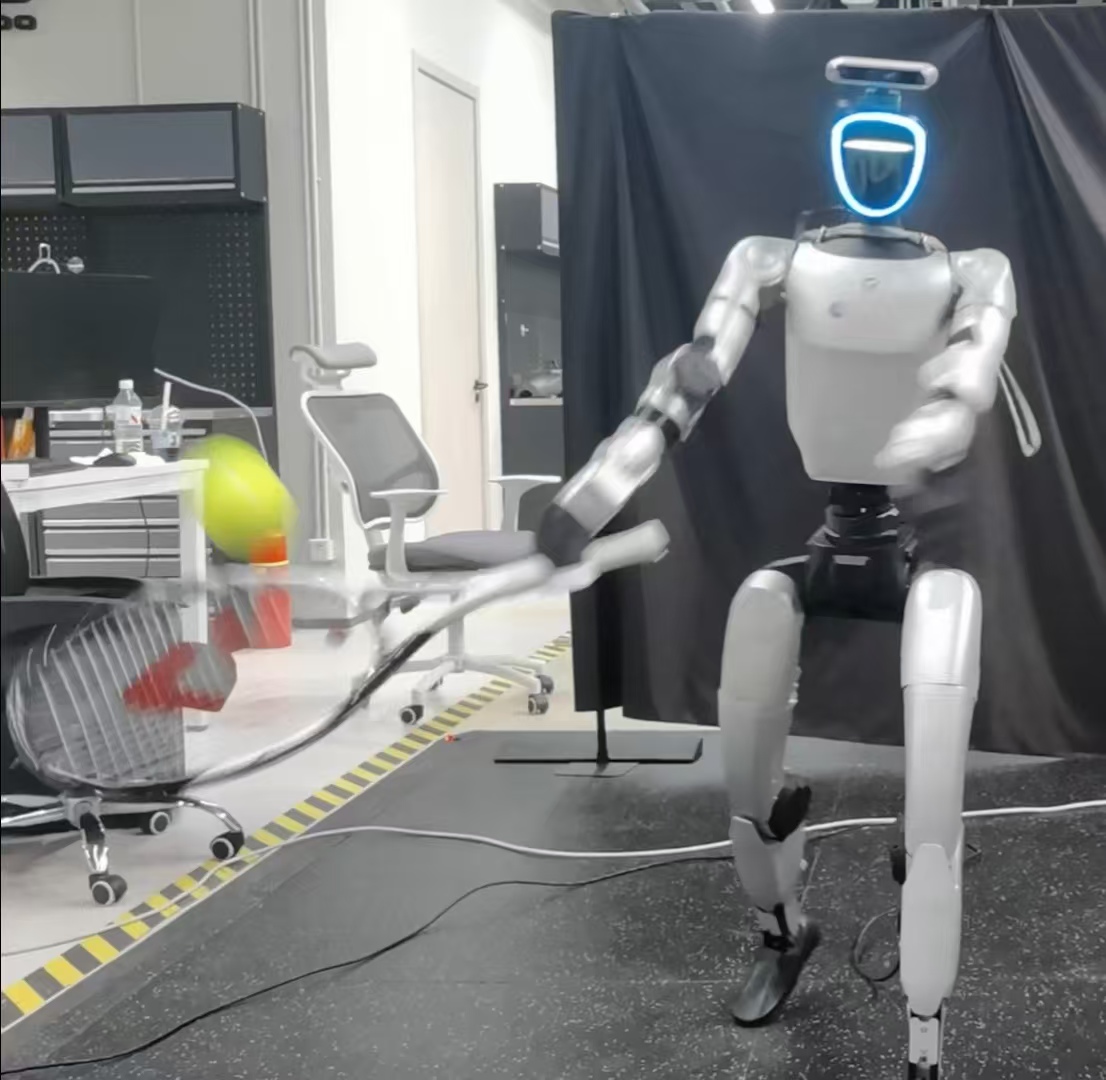}
  \end{subfigure}\hfill
  \begin{subfigure}[t]{0.24\textwidth}
    \centering
    \includegraphics[width=\linewidth,height=3.5cm]{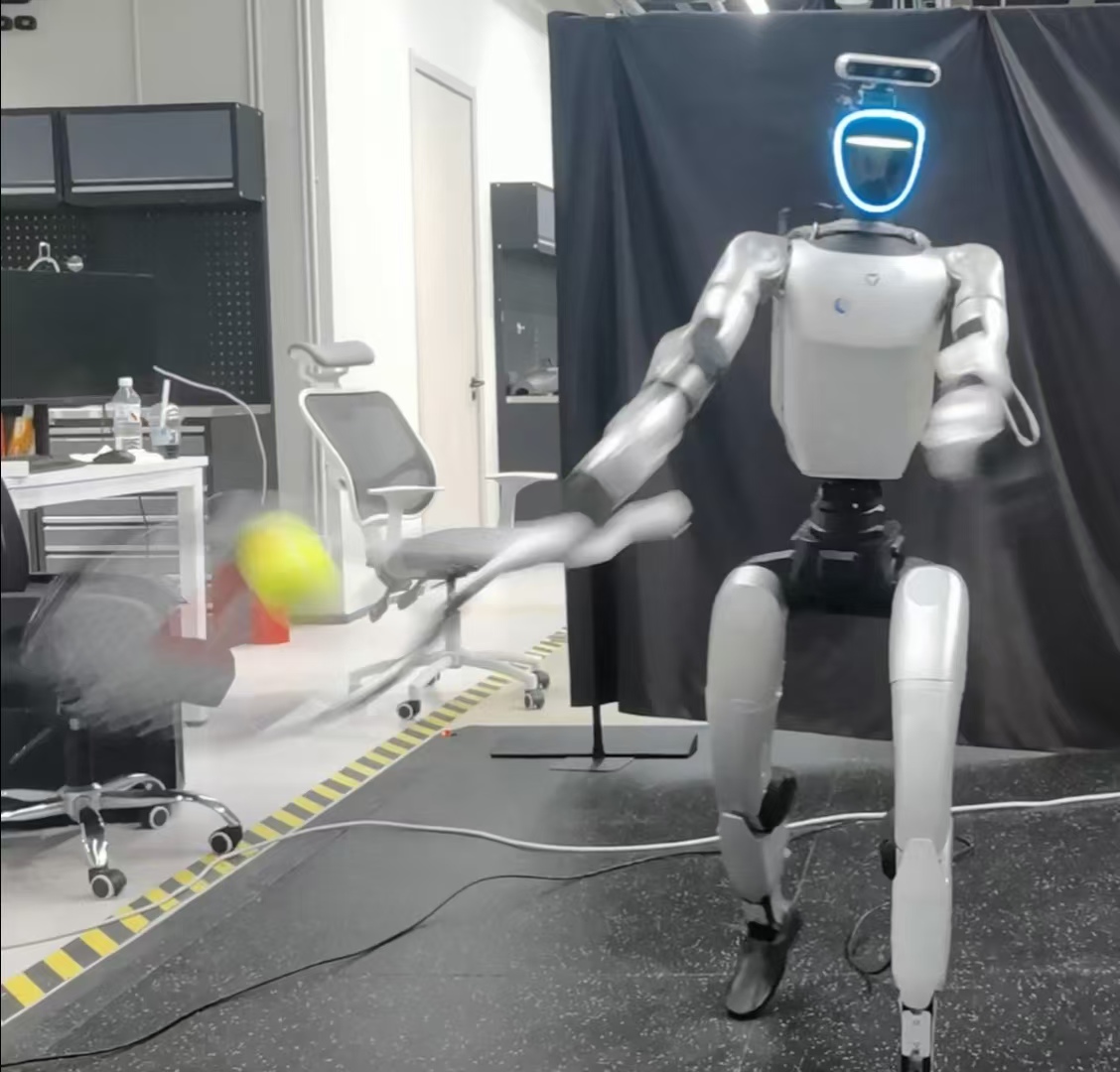}
  \end{subfigure}\hfill
  \begin{subfigure}[t]{0.24\textwidth}
    \centering
    \includegraphics[width=\linewidth,height=3.5cm]{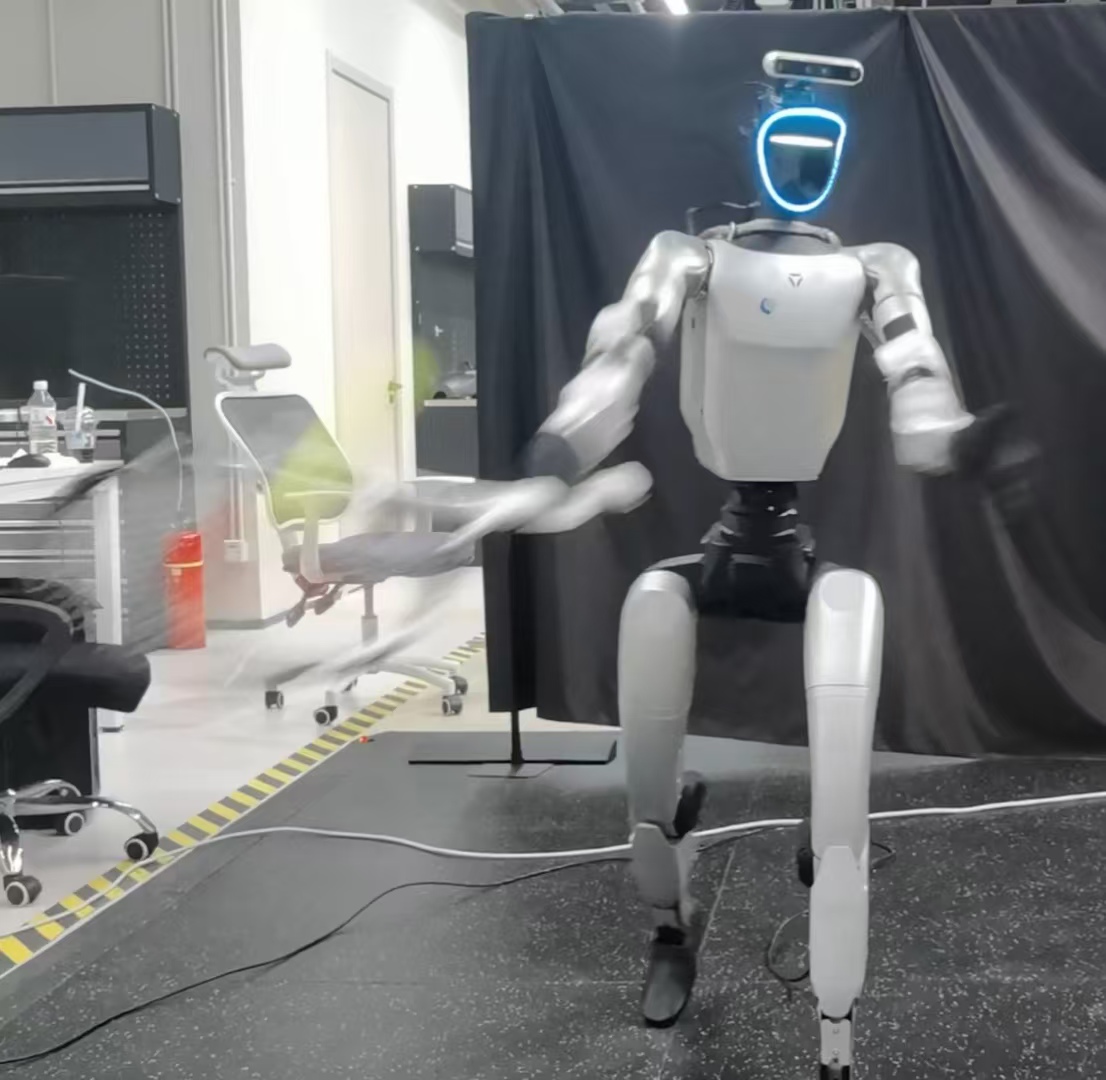}
  \end{subfigure}\hfill
  \begin{subfigure}[t]{0.24\textwidth}
    \centering
    \includegraphics[width=\linewidth,height=3.5cm]{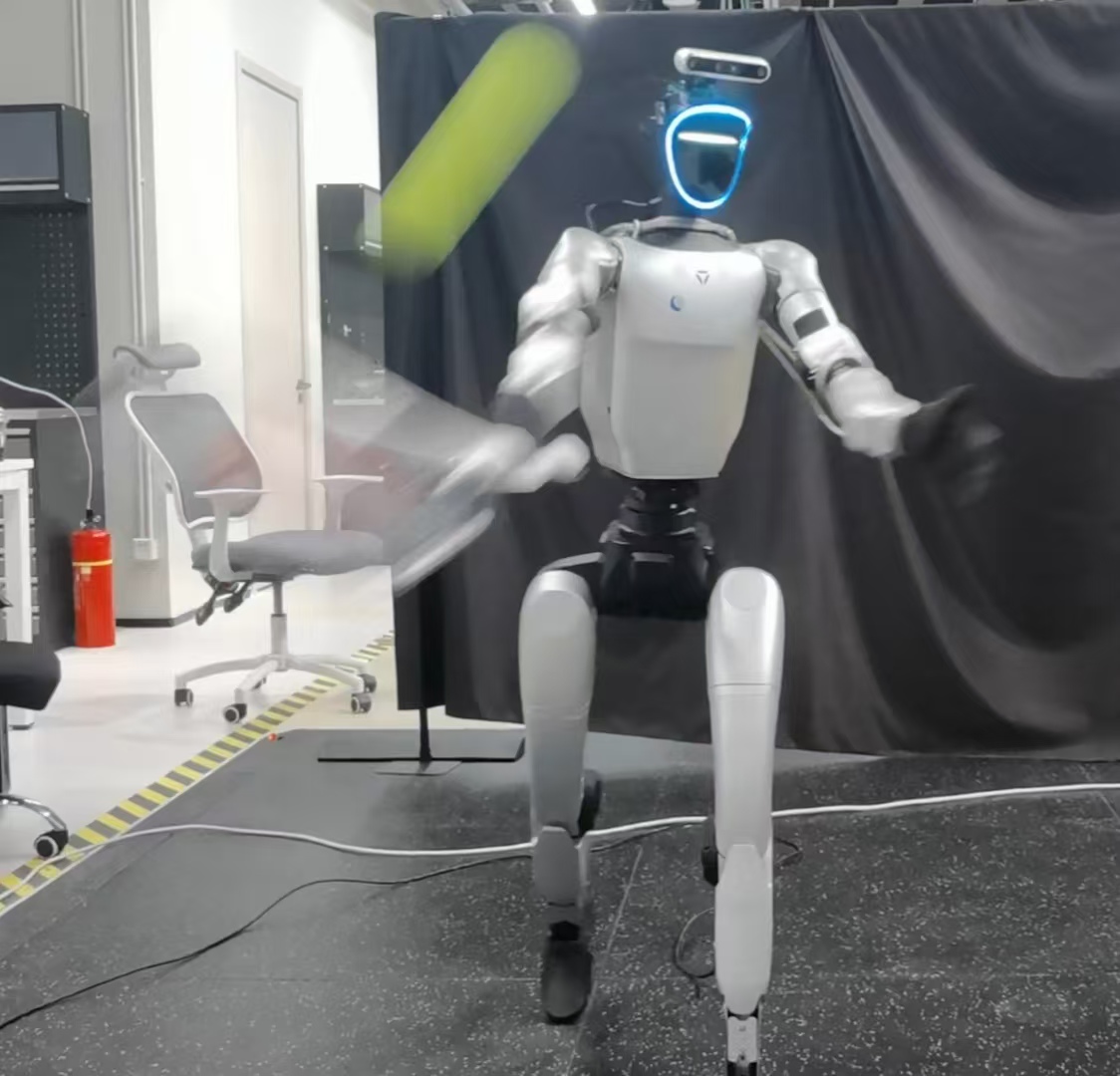}
  \end{subfigure}
  \caption{Representative sequence of vision-based tennis hitting on the humanoid robot.}
  \label{fig:t}
\end{figure*}

\begin{figure*}[t]
  \centering
  \begin{subfigure}[t]{0.24\textwidth}
    \centering
    \includegraphics[width=\linewidth,height=3.5cm]{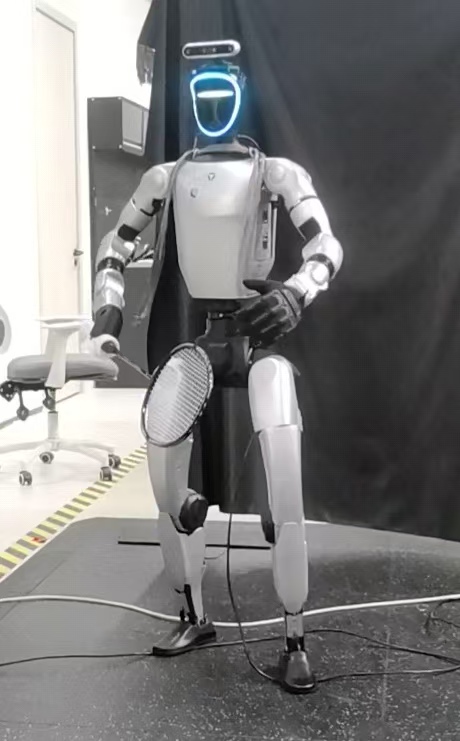}
  \end{subfigure}\hfill
  \begin{subfigure}[t]{0.24\textwidth}
    \centering
    \includegraphics[width=\linewidth,height=3.5cm]{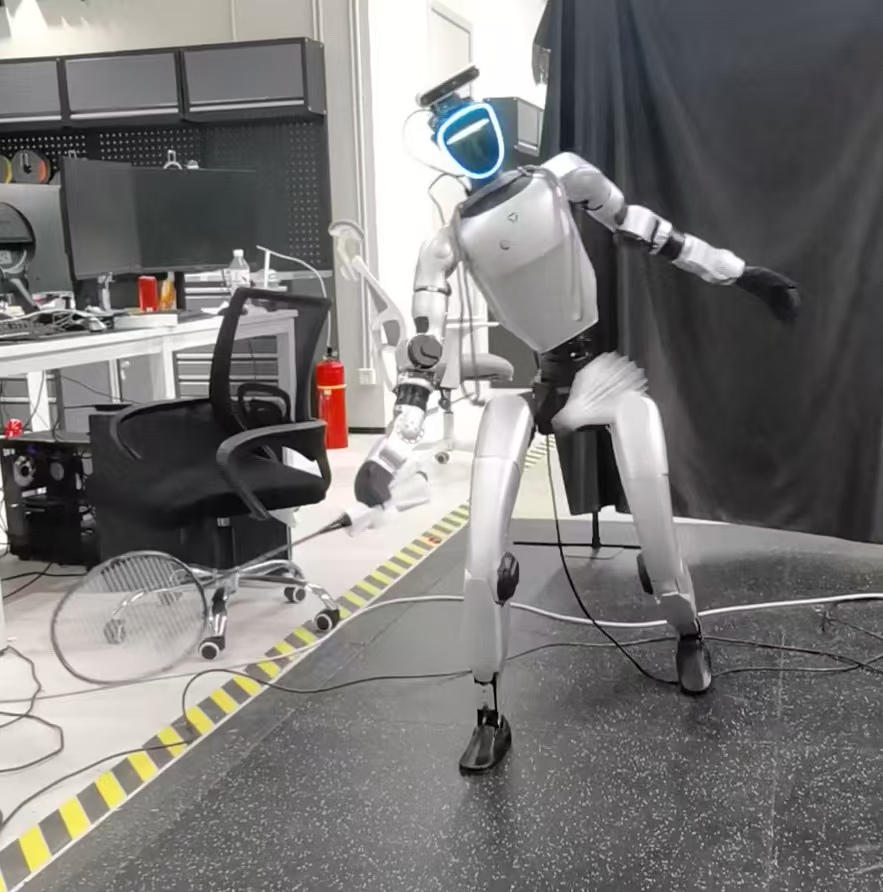}
  \end{subfigure}\hfill
  \begin{subfigure}[t]{0.24\textwidth}
    \centering
    \includegraphics[width=\linewidth,height=3.5cm]{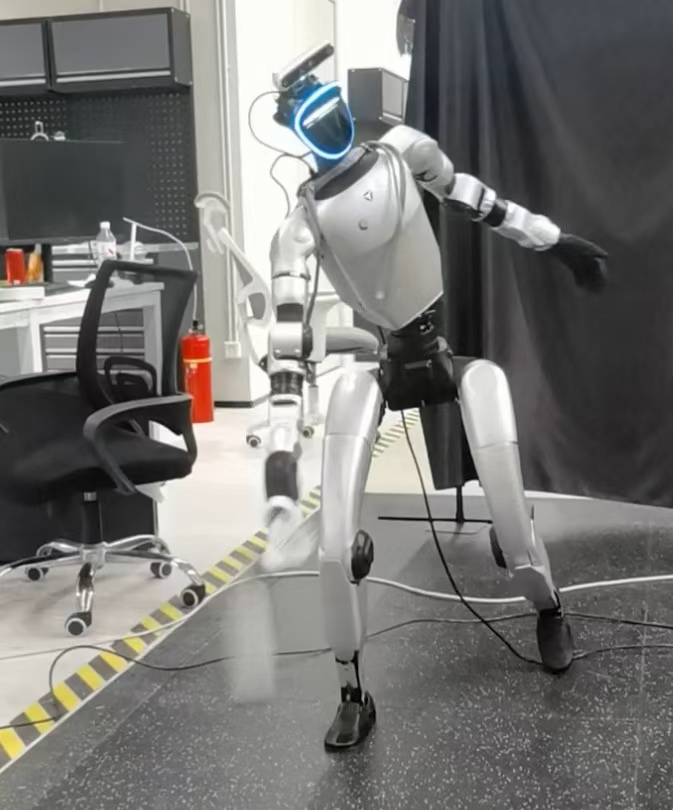}
  \end{subfigure}\hfill
  \begin{subfigure}[t]{0.24\textwidth}
    \centering
    \includegraphics[width=\linewidth,height=3.5cm]{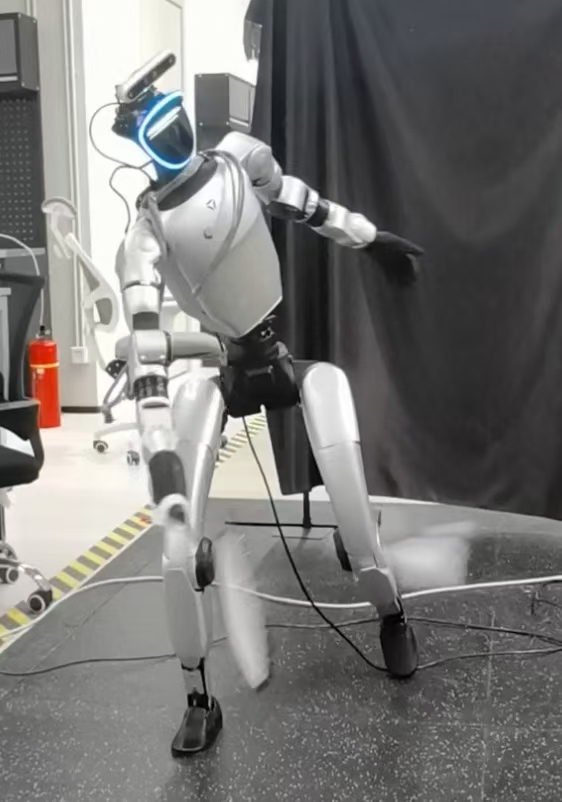}
  \end{subfigure}
  \caption{Representative sequence of vision-based badminton hitting on the humanoid robot.}
  \label{fig:b}
\end{figure*}

\subsection{Task-Space Target Generation}

The active interception plan $\Pi_t$ is converted into a set of structured task-space targets rather than joint-level commands. These targets define \emph{what} motion should be realized at the task level, while the low-level realization of locomotion, balance, and whole-body coordination is handled by SONIC.

The head target pose, denoted by $\mathbf{T}_{\mathrm{head}}\in SE(3)$, is generated online to maintain visual attention on the incoming projectile. Specifically, the desired head orientation is chosen such that the nominal head forward axis aligns with the current projectile direction in the robot frame. This allows the robot to continuously observe the projectile during both the approach and swing phases.

The wrist targets are generated from a phase-dependent strike template. For the non-dominant arm, we maintain a relatively stable support pose to avoid unnecessary upper-body disturbance. For the racket arm, the target wrist pose is computed from the current swing mode $m_t$, the desired contact point $\mathbf{p}_{\mathrm{hit}}$, the desired base placement $\mathbf{p}_{\mathrm{base}}^{\mathrm{des}}$, and the desired swing direction $\mathbf{d}_{\mathrm{swing}}$. Let
\[
\mathbf{T}_{\mathrm{wrist}}^{\mathrm{ready}},\quad
\mathbf{T}_{\mathrm{wrist}}^{\mathrm{cock}},\quad
\mathbf{T}_{\mathrm{wrist}}^{\mathrm{contact}},\quad
\mathbf{T}_{\mathrm{wrist}}^{\mathrm{follow}}
\in SE(3)
\]
denote the nominal racket-wrist poses for the ready, backswing, contact, and follow-through phases, respectively. The commanded wrist poses for the left and right wrists are denoted by $\mathbf{T}_{\mathrm{wrist}}^{L}\in SE(3)$ and $\mathbf{T}_{\mathrm{wrist}}^{R}\in SE(3)$, respectively. In practice, the active wrist command is obtained by phase-dependent interpolation between the nominal templates, with the contact pose additionally aligned to the predicted interception geometry. This yields a smooth upper-body motion prior while still allowing the whole-body controller to compensate for balance and timing errors.

In addition to the head and wrist targets, the planner outputs a desired base height and a navigation command. The desired base height, denoted by $h_{\mathrm{base}}$, is adjusted to keep the hitting posture within a comfortable arm workspace. The navigation command, denoted by $\mathbf{u}_{\mathrm{nav}}$, drives the base toward the target placement $\mathbf{p}_{\mathrm{base}}^{\mathrm{des}}$ with the desired heading $\psi_{\mathrm{des}}$. Together, these quantities define a compact high-level motion intent:
\begin{equation}
\mathbf{u}_{\mathrm{plan}}
=
\bigl(
\mathbf{T}_{\mathrm{head}},
\mathbf{T}_{\mathrm{wrist}}^{L},
\mathbf{T}_{\mathrm{wrist}}^{R},
h_{\mathrm{base}},
\mathbf{u}_{\mathrm{nav}}
\bigr),
\label{eq:planner_command}
\end{equation}
where $\mathbf{T}_{\mathrm{head}}$ and $\mathbf{T}_{\mathrm{wrist}}^{L/R}$ are $SE(3)$ targets for the head and wrists, $h_{\mathrm{base}}$ is the desired base height, and $\mathbf{u}_{\mathrm{nav}}$ is the planar navigation command for root motion.

\subsection{Execution Interface to SONIC}

To decouple perception from execution, the visual estimation stack is exposed to the planner through a unified projectile-state interface. Internally, this module encapsulates RGB-D acquisition, 2D detection, temporal tracking, 3D localization, EKF filtering, and short-horizon trajectory prediction. Externally, however, the planner consumes only a compact projectile-state representation
\[
\mathcal{B}_t=(\mathbf{p}_t,\mathbf{v}_t,c_t),
\]
where $\mathbf{p}_t\in\mathbb{R}^3$ and $\mathbf{v}_t\in\mathbb{R}^3$ are the filtered projectile position and velocity estimates, and $c_t\in[0,1]$ denotes the current confidence. This abstraction hides perception-specific details from the planning layer and allows the planner to operate on a clean dynamical state interface.

Given $\mathcal{B}_t$, the interception planner computes the active plan $\Pi_t$ in~(\ref{eq:intercept_plan}) and converts it into the task-space command $\mathbf{u}_{\mathrm{plan}}$ in~(\ref{eq:planner_command}). The planner output is then streamed to SONIC as structured motion commands. These commands include the locomotion mode, planar movement and facing commands, desired speed, base height, and the $SE(3)$ targets for the head and wrists. SONIC interprets them as hybrid kinematic commands and realizes them through its whole-body tracking policy.

Importantly, this interface does not require SONIC to reason about detection, tracking, or projectile-state estimation; it only consumes compact task-level motion targets. Conversely, the planner does not solve inverse kinematics or whole-body balance explicitly. The resulting architecture forms a clean perception--planning--control stack: the perception module produces an uncertainty-aware projectile state, the planner converts it into an interception intent, and SONIC realizes that intent as dynamically feasible whole-body motion.

\section{Experiments}

We evaluate \emph{CyboRacket} in both simulation and the real world, focusing on integrated hitting performance in simulation and transfer to Unitree G1. Representative motion sequences are shown in Fig.~\ref{fig:s} for simulation, and in Fig.~\ref{fig:t} and Fig.~\ref{fig:b} for real-world tennis and badminton hitting, respectively.

\subsection{Simulation Experiments in MuJoCo}

We evaluate the complete \emph{CyboRacket} pipeline in a MuJoCo simulation environment that includes the humanoid robot, racket geometry, tennis ball dynamics, and court layout. In each rollout, the robot starts from a nominal ready stance, while the incoming object is launched under varying conditions. In each trial, the robot perceives the incoming object, predicts its future trajectory, plans an interception target, and executes the corresponding whole-body motion. A representative simulation sequence is shown in Fig.~\ref{fig:s}. The evaluation is conducted over 20 simulated trials. The proposed system achieves a hit rate of 75\% and a return rate of 70\%.

\subsection{Real-World Experiments}

We evaluate \emph{CyboRacket} on the Unitree G1 humanoid robot, equipped with a racket mounted on the right arm and an external Intel RealSense D455 RGB-D camera mounted on the head for onboard ball perception. The camera streams synchronized RGB-D observations at 90\,Hz. In the tennis setting, incoming balls are launched by a ball machine under varying speeds and directions. The complete \emph{CyboRacket} pipeline is deployed in a vision-based tennis-hitting task, where the robot starts from a ready pose, perceives the incoming tennis ball using the onboard RGB-D camera, predicts its future trajectory, plans an interception target, and executes the corresponding whole-body motion through SONIC. A representative real-world tennis-hitting sequence is shown in Fig.~\ref{fig:t}. These results demonstrate that the proposed perception-to-action pipeline can be transferred from simulation to the real humanoid platform using purely onboard sensing.

To further illustrate the generality of the proposed framework beyond tennis, we additionally present a vision-based badminton-hitting sequence on the humanoid robot, as shown in Fig.~\ref{fig:b}. This result further suggests that the proposed perception-to-action pipeline can be extended to broader humanoid racket-sport interaction scenarios.

\section{Conclusion}

In this work, we have presented \emph{CyboRacket}, a hierarchical perception-to-action framework for humanoid racket sports that has combined onboard visual perception, physics-based trajectory prediction, and large-scale pre-trained whole-body control. We have instantiated the framework on the Unitree G1 humanoid robot and have validated it in a vision-based tennis-hitting task using purely onboard sensing. The experimental results have shown that the proposed system has supported real-time object tracking, trajectory prediction, and successful striking through coordinated whole-body execution. These findings have demonstrated the potential of integrating closed-loop perception and planning with pre-trained humanoid motion control for dynamic racket-sport interaction, and have provided a practical foundation for extending such capabilities to broader humanoid athletic skills.

\ifCLASSOPTIONcaptionsoff
  \newpage
\fi

\bibliographystyle{IEEEtran}
\bibliography{references} 

@inproceedings{long2025learning,
  title={Learning humanoid locomotion with perceptive internal model},
  author={Long, Jing and Ren, Jiawei and Shi, Ming and Wang, Zheng and Huang, Tenglong and Luo, Ping and Pang, Jiangmiao},
  booktitle={2025 IEEE International Conference on Robotics and Automation (ICRA)},
  year={2025},
  organization={IEEE}
}

@article{zhang2025hub,
  title={Hub: Learning extreme humanoid balance},
  author={Zhang, Tairan and Zheng, Boyuan and Nair, Ram and Hu, Yuming and Wang, Yen-Jui and Chen, Guanbo and Lin, Fan and Li, Jiawei and Hong, Chenyu and Sreenath, Koushil and others},
  journal={arXiv preprint arXiv:2505.07294},
  year={2025},
  archivePrefix={arXiv},
  eprint={2505.07294},
  primaryClass={cs.RO}
}

@inproceedings{cheng2024expressive,
  title={Expressive whole-body control for humanoid robots},
  author={Cheng, Xuxin and Ji, Yuxuan and Chen, Jie and Yang, Ruiqi and Yang, Guowei and Wang, Xiaolong},
  booktitle={20th Robotics: Science and Systems (RSS 2024)},
  year={2024},
  organization={MIT Press Journals}
}

@inproceedings{tebbe2021sample,
  title={Sample-efficient reinforcement learning in robotic table tennis},
  author={Tebbe, Jonas and Krauch, Lukas and Gao, Yuman and Zell, Andreas},
  booktitle={2021 IEEE International Conference on Robotics and Automation (ICRA)},
  pages={4171--4178},
  year={2021},
  organization={IEEE}
}

@inproceedings{dambrosio2023robotic,
  title={Robotic Table Tennis: A Case Study into a High Speed Learning System},
  author={D'Ambrosio, Davide B. and Jaitly, Navdeep and Sindhwani, Vikas and Oslund, Ken and Xu, Peng and Lazic, Nevena and Shankar, Adithya and Abelian, Jose and Coumans, Erwin and others},
  booktitle={Proceedings of Robotics: Science and Systems (RSS)},
  address={Daegu, Republic of Korea},
  year={2023}
}

@inproceedings{dambrosio2025achieving,
  title={Achieving human level competitive robot table tennis},
  author={D'Ambrosio, Davide B. and Abeyruwan, Saminda and Graesser, Laura and Iscen, A. and Ben Amor, Heni and Bewley, Alex and Reed, Brian J. and Reymann, Katharina and Takayama, Leila and Tassa, Yuval and others},
  booktitle={2025 IEEE International Conference on Robotics and Automation (ICRA)},
  year={2025},
  organization={IEEE}
}

@inproceedings{nguyen2025high,
  title={High speed robotic table tennis swinging using lightweight hardware with model predictive control},
  author={Nguyen, Dac and Cancio, Kevin D. and Kim, Sangbae},
  booktitle={2025 IEEE International Conference on Robotics and Automation (ICRA)},
  year={2025},
  organization={IEEE}
}

@article{su2025hitter,
  title={HITTER: A HumanoId Table TEnnis Robot via Hierarchical Planning and Learning},
  author={Su, Zhi and Zhang, Bike and Rahmanian, Nima and Gao, Yuman and Liao, Qiayuan and Regan, Caitlin and Sreenath, Koushil and Sastry, S. Shankar},
  journal={arXiv preprint arXiv:2508.21043},
  year={2025},
  archivePrefix={arXiv},
  eprint={2508.21043},
  primaryClass={cs.RO}
}

@article{ma2025learning,
  title={Learning coordinated badminton skills for legged manipulators},
  author={Ma, Yuntao and Cramariuc, Andrei and Farshidian, Farbod and Hutter, Marco},
  journal={Science Robotics},
  volume={10},
  number={102},
  pages={eadu3922},
  year={2025},
  publisher={American Association for the Advancement of Science}
}

@article{luo2025sonic,
  title={SONIC: Supersizing Motion Tracking for Natural Humanoid Whole-Body Control},
  author={Luo, Zhengyi and Yuan, Ye and Wang, Tingwu and Li, Chenran and Chen, Sirui and Casta{\\~n}eda, Fernando and Cao, Zi-Ang and Li, Jiefeng and Minor, David and Ben, Qingwei and Da, Xingye and Ding, Runyu and Hogg, Cyrus and Song, Lina and Lim, Edy and Jeong, Eugene and He, Tairan and Xue, Haoru and Xiao, Wenli and Wang, Zi and Yuen, Simon and Kautz, Jan and Chang, Yan and Iqbal, Umar and Fan, Linxi "Jim" and Zhu, Yuke},
  journal={arXiv preprint arXiv:2511.07820},
  year={2025},
  archivePrefix={arXiv},
  eprint={2511.07820},
  primaryClass={cs.RO}
}

@inproceedings{mulling2010biomimetic,
  title={A biomimetic approach to robot table tennis},
  author={M{\"u}lling, Katharina and Kober, Jens and Peters, Jan},
  booktitle={2010 IEEE/RSJ International Conference on Intelligent Robots and Systems (IROS)},
  pages={1921--1926},
  year={2010},
  organization={IEEE},
  note={Also published in \textit{Adaptive Behavior}, vol. 19, no. 5, pp. 359--376, 2011}
}

@article{mulling2013learning,
  title={Learning to select and generalize striking movements in robot table tennis},
  author={M{\"u}lling, Katharina and Kober, Jens and Kroemer, Oliver and Peters, Jan},
  journal={The International Journal of Robotics Research (IJRR)},
  volume={32},
  number={3},
  pages={263--279},
  year={2013},
  publisher={SAGE Publications}
}

@inproceedings{wang2011learning,
  title={Learning anticipation policies for robot table tennis},
  author={Wang, Zhikun and Lampert, Christoph H. and M{\"u}lling, Katharina and Sch{\"o}lkopf, Bernhard and Peters, Jan},
  booktitle={2011 IEEE/RSJ International Conference on Intelligent Robots and Systems (IROS)},
  pages={332--337},
  year={2011},
  organization={IEEE}
}

@article{wang2017anticipatory,
  title={Anticipatory action selection for human--robot table tennis},
  author={Wang, Zhikun and Boularias, Abdeslam and M{\"u}lling, Katharina and Sch{\"o}lkopf, Bernhard and Peters, Jan},
  journal={Artificial Intelligence},
  volume={247},
  pages={399--414},
  year={2017},
  publisher={Elsevier},
  note={Special Issue on AI and Robotics}
}

@article{zaidi2023athletic,
  title={Athletic mobile manipulator system for robotic wheelchair tennis},
  author={Zaidi, Zak and Martin, Dominic and Belles, Nicolas and Zakharov, Vladimir and Krishna, Akhil and Lee, Ken Ming and Wagstaff, Peter and Naik, Saranjeet and Sklar, Michael and Choi, Sehoon and others},
  journal={IEEE Robotics and Automation Letters (RA-L)},
  volume={8},
  number={4},
  pages={2245--2252},
  year={2023},
  publisher={IEEE}
}

@article{xiong2012impedance,
  title={Impedance control and its effects on a humanoid robot playing table tennis},
  author={Xiong, Rong and Sun, Yue and Zhu, Qiuguo and Wu, Jianguo and Chu, Jian},
  journal={International Journal of Advanced Robotic Systems (IJARS)},
  volume={9},
  number={5},
  pages={178},
  year={2012},
  publisher={SAGE Publications}
}

@article{peng2018deepmimic,
  title={DeepMimic: Example-guided deep reinforcement learning of physics-based character skills},
  author={Peng, Xue Bin and Abbeel, Pieter and Levine, Sergey and van de Panne, Michiel},
  journal={ACM Transactions on Graphics (TOG)},
  volume={37},
  number={4},
  pages={1--14},
  year={2018},
  month={July},
  publisher={ACM},
  note={Proceedings of SIGGRAPH 2018}
}

@article{peng2021amp,
  title={AMP: Adversarial motion priors for stylized physics-based character control},
  author={Peng, Xue Bin and Zhou, Zhaoyu and Luo, Stephen and van de Panne, Michiel},
  journal={ACM Transactions on Graphics (TOG)},
  volume={40},
  number={4},
  pages={1--20},
  year={2021},
  month={July},
  publisher={ACM},
  note={Proceedings of SIGGRAPH 2021}
}

@inproceedings{he2024omnih2o,
  title={OmniH2O: Universal and dexterous human-to-humanoid whole-body teleoperation and learning},
  author={He, Tairan and Luo, Zhengyi and Xiao, Wenli and Zhang, Chong and Zhang, Weinan and Kitani, Kris and Liu, Changliu and Shi, Guanya},
  booktitle={8th Annual Conference on Robot Learning (CoRL)},
  year={2024},
  organization={PMLR}
}

@article{zhang2025falcon,
  title={Falcon: Learning force-adaptive humanoid loco-manipulation},
  author={Zhang, Yandong and Yuan, Ye and Gurunath, Prabhat and He, Tairan and Omidshafiei, Shayegan and Aghamohammadi, Aria and Vazquez-Chanlatte, Marlos and Pedersen, Leila and Shi, Guanya},
  journal={arXiv preprint arXiv:2505.06776},
  year={2025},
  archivePrefix={arXiv},
  eprint={2505.06776},
  primaryClass={cs.RO}
}

@article{liao2025beyondmimic,
  title={BeyondMimic: From motion tracking to versatile humanoid control via guided diffusion},
  author={Liao, Qiayuan and Truong, Takara E. and Huang, Xiaoyu and Tevet, Guy and Sreenath, Koushil and Liu, C. Karen},
  journal={arXiv preprint arXiv:2508.08241},
  year={2025},
  archivePrefix={arXiv},
  eprint={2508.08241},
  primaryClass={cs.RO}
}

@article{chen2025gmt,
  title={GMT: General motion tracking for humanoid whole-body control},
  author={Chen, Zixuan and Ji, Mazeyu and Cheng, Xuxin and Peng, Xuanbin and Peng, Xue Bin and Wang, Xiaolong},
  journal={arXiv preprint arXiv:2506.14770},
  year={2025},
  archivePrefix={arXiv},
  eprint={2506.14770},
  primaryClass={cs.RO}
}

@inproceedings{ze2025twist,
  title={TWIST: Teleoperated Whole-Body Imitation System},
  author={Ze, Yanjie and Chen, Zixuan and Araujo, Joao Pedro and Cao, Zi-ang and Peng, Xue Bin and Wu, Jiajun and Liu, Karen},
  booktitle={Proceedings of The 9th Conference on Robot Learning},
  year={2025},
  volume={305},
  pages={2143--2154},
  publisher={PMLR}
}

@article{zhang2025track,
  title={Track any motions under any disturbances},
  author={Zhang, Zhikai and Guo, Jun and Chen, Chao and Wang, Jilong and Lin, Chenghuai and Lian, Yunrui and Xue, Han and Wang, Zhenrong and Liu, Maoqi and Liu, Huaping and others},
  journal={arXiv preprint arXiv:2509.13833},
  year={2025},
  archivePrefix={arXiv},
  eprint={2509.13833},
  primaryClass={cs.RO}
}

@article{bjorck2025groot,
  title={GR00T N1: An open foundation model for generalist humanoid robots},
  author={Bjorck, Johan and Casta{\~n}eda, Fernando and Cherniadev, Nikita and Da, Xingye and Ding, Runyu and Fan, Linxi "Jim" and Fang, Yu and Fox, Dieter and Hu, Fengyuan and Huang, Spencer and others},
  journal={arXiv preprint arXiv:2503.14734},
  year={2025},
  archivePrefix={arXiv},
  eprint={2503.14734},
  primaryClass={cs.RO}
}

\end{document}